\documentclass[10pt,twocolumn]{article}
\IfFileExists{lmodern.sty}{\usepackage{lmodern}}{}
\usepackage{amssymb,amsmath}
\usepackage{ifxetex,ifluatex}
\IfFileExists{fixltx2e.sty}{\usepackage{fixltx2e}}{}
\ifnum0\ifxetex1\fi\ifluatex1\fi=0
  \usepackage[T1]{fontenc}
  \usepackage[utf8]{inputenc}
\else
  \ifxetex
    \usepackage{mathspec}
    \usepackage{xltxtra,xunicode}
  \else
    \usepackage{fontspec}
  \fi
  \defaultfontfeatures{Mapping=tex-text,Scale=MatchLowercase}
\fi
\IfFileExists{upquote.sty}{\usepackage{upquote}}{}
\IfFileExists{microtype.sty}{\usepackage{microtype}}{}
\usepackage{fvextra}
\DefineVerbatimEnvironment{verbatim}{Verbatim}{breaklines=true,breakanywhere=true,fontsize=\footnotesize}
\IfFileExists{geometry.sty}{\usepackage[letterpaper,top=0.8in,bottom=0.9in,left=0.75in,right=0.75in]{geometry}}{}
\usepackage{array,tabularx,booktabs}
\usepackage{graphicx}
\ifxetex
  \usepackage[setpagesize=false,unicode=false,xetex]{hyperref}
\else
  \usepackage[unicode=true]{hyperref}
\fi
\hypersetup{breaklinks=true,
            bookmarks=true,
            pdfauthor={Luc Debaupte, Tyler Baumgartner, Brandon Tai, Candice Fan, Bill Wang, Yi Zhong},
            pdftitle={VocalAffectBench: Evaluating Vocal Emotion Recognition in AI Audio Models},
            colorlinks=true,
            citecolor=blue,
            urlcolor=blue,
            linkcolor=magenta,
            pdfborder={0 0 0}}
\newcolumntype{L}{>{\raggedright\arraybackslash}X}

\title{VocalAffectBench: Evaluating Vocal Emotion Recognition in AI Audio Models}
\author{Luc Debaupte, Tyler Baumgartner, Brandon Tai,\\
Candice Fan, Bill Wang, and Yi Zhong\\
Besimple AI, San Mateo, CA\\
\texttt{\{luc, yi\}@besimple.ai}}
\date{}

\begin{document}

\maketitle

\begin{abstract}
Voice products increasingly need affective cues that are present in
speech but absent from transcripts. We introduce
\textbf{VocalAffectBench}, a public, test-only benchmark for evaluating
whether AI audio models can identify expressed vocal emotion from raw
audio. The benchmark contains 280 human-recorded English WAV clips from
52 speaker accounts totaling 2.32 hours across seven labels:
\texttt{angry}, \texttt{disgusted}, \texttt{fearful}, \texttt{happy},
\texttt{neutral}, \texttt{sad}, and \texttt{surprised}, with 40 clips
per class. All baselines are evaluated from audio alone, without
transcripts or contextual metadata.

Across six released baselines, average accuracy is 35.1\%.
The strongest baseline, \texttt{gemini\_3\_5\_flash}, reaches 44.3\% on
the seven-way task, above the 14.3\% random baseline but far from robust
emotion recognition. A secondary valence-bucket analysis maps labels
into positive, neutral, and negative classes, excluding
\texttt{surprised} because its valence is ambiguous. Aggregate accuracy
under this coarser view is 49.2\%. Performance is highly uneven across
classes. By recall, neutral is identified most reliably at 76.2\%
averaged across baselines, while surprised and fearful reach only 15.0\%
each.
These results show that the evaluated baselines can extract some
affective signal from speech, but discrete expressed-emotion recognition
remains fragile, especially for non-neutral emotions that are often most
important in voice agent workflows.
\end{abstract}

\section{Introduction}\label{introduction}

Voice agents increasingly mediate interactions where tone, pacing,
pauses, and intensity can change the meaning of otherwise identical
words. A customer who says ``that's fine'' may be conceding, joking,
withdrawing, or signaling frustration. A transcript preserves the phrase,
but not the delivery that makes the interaction interpretable.

Automatic speech recognition (ASR) transcripts are often insufficient
for this purpose. Text-only sentiment analysis can identify emotionally
loaded words, but it answers a different question from speech emotion
recognition. Voice agents that apply affect analysis only after
transcription may therefore measure lexical sentiment rather than
expressed vocal emotion.

Existing affective-speech resources have advanced the field through
reusable corpora and shared evaluation tasks (Busso et al., 2008;
Livingstone and Russo, 2018; Cao et al., 2014; Schuller et al., 2018).
However, model comparisons are often difficult to interpret when task
design and output spaces differ. For production teams choosing among
hosted audio models, a compact raw-audio benchmark with public
predictions and a fixed label set can be more actionable than a large
training corpus.

VocalAffectBench addresses this gap with a small, auditable test set for
single-label expressed vocal emotion in English speech. The benchmark is
test-only, with equal class counts in the released set. Every baseline
is scored on the same 280 clips under the same input protocol and label
mapping policy, making the comparison easy to inspect and repeat.

\section{Methods}\label{methods}

\subsection{Benchmark Design}\label{benchmark-design}

VocalAffectBench is a test-only benchmark for expressed vocal emotion
recognition. Each item contains a human-recorded English audio clip, a
single verified target emotion label, and basic recording metadata. The
released audio is 16 kHz mono WAV.

The target is \emph{expressed} vocal emotion. During collection,
speakers received script text and performed it under a specific assigned
emotion. The script text was written to be compatible with multiple
assigned emotions, so the label comes from the requested delivery rather
than from emotional words in the script. We retained clips only when the
delivered performance matched the assigned label during human review. The
benchmark therefore measures whether a model identifies the emotion
expressed in the recording, not whether it can infer what the speaker
privately felt.

\begin{table}[t]\centering\footnotesize
  \caption{Dataset composition. Each label has 40 clips.}\label{tab:dataset-composition}
  \begin{tabular}{lrrr}
    \toprule
    Label & Clips & Minutes & Mean sec. \\
    \midrule
    \texttt{angry} & 40 & 17.1 & 25.7 \\
    \texttt{disgusted} & 40 & 19.0 & 28.5 \\
    \texttt{fearful} & 40 & 17.5 & 26.3 \\
    \texttt{happy} & 40 & 22.8 & 34.2 \\
    \texttt{neutral} & 40 & 15.6 & 23.3 \\
    \texttt{sad} & 40 & 26.0 & 38.9 \\
    \texttt{surprised} & 40 & 21.0 & 31.6 \\
    \midrule
    Total & 280 & 138.9 & 29.8 \\
    \bottomrule
  \end{tabular}
\end{table}

\subsection{Collection and Curation}\label{collection-and-curation}

Audio clips were collected from human speakers performing emotional
speech between May 15 and June 2, 2026. Collection used a same-script
protocol. For each task, speakers received one short generated script
and recorded it in three assigned emotions. The script was generated so
the same words could plausibly be delivered in each of those emotions
rather than lexically forcing one label.

After collection, one reviewer checked each clip for audio quality and
whether the expressed delivery matched the requested emotion. Clips were
excluded if audio quality was insufficient or if the performed emotion
was incorrect or unclear. The reviewed pool available for benchmark
construction contained 563 clips from 68 unique speakers.

We constructed the public benchmark by selecting 40 reviewed clips for
each of the seven emotion labels. All public clips are single-speaker English recordings.
Each clip has a General American accent and no background noise. We exported the released
audio as 16 kHz mono WAV, with no noise reduction,
normalization, or enhancement.

\subsection{Evaluation Protocol}\label{evaluation-protocol}

The six baselines are evaluated under a raw-audio-input protocol. At
inference time, prompted audio models receive the audio file and a fixed
instruction listing the allowed labels:

\begin{quote}\footnotesize
Choose exactly one primary expressed emotion from the allowed label set.
Base your answer only on the expressed vocal tone, prosody, pace,
intensity, pauses, and wording. Do not infer the speaker's private
internal state.
\end{quote}

No transcript or contextual metadata is provided to the model. For
provider emotion or prosody endpoints that do not accept arbitrary
prompts, the evaluation submits the audio file to the endpoint with its
default settings. The endpoint returns its native affect labels or
scores, which are then mapped to the benchmark label set before scoring.
The main evaluation uses a closed seven-label classification target:
\begin{center}
\small
\begin{tabular}{llll}
\texttt{angry} & \texttt{disgusted} & \texttt{fearful} & \texttt{happy} \\
\texttt{neutral} & \texttt{sad} & \texttt{surprised}
\end{tabular}
\end{center}

This target set is intentionally conventional rather than exhaustive.
The six non-neutral labels follow Ekman's widely used basic-emotion taxonomy
(Ekman, 1992). We include \texttt{neutral} because prior
speech-emotion benchmarks use it as a control or non-emotional class,
including RAVDESS (Livingstone and Russo, 2018). It also reflects a
practical requirement for deployed voice agents, which often need to
separate marked affect from ordinary delivery.
This choice does not assume that all emotion is reducible to seven
universal categories. It provides a compact, familiar, auditable label
space for comparing audio models.

Provider outputs may use native affect labels or scores. For scoring,
each returned label is mapped to the seven-label benchmark set using the
mapping implemented in the released evaluation script. For example,
calm-style outputs map to \texttt{neutral}, and anxiety-style outputs
map to \texttt{fearful}. The released \texttt{predictions.csv} preserves
both the raw provider output and the normalized \texttt{mapped\_label}.
The full mapping is reported in the appendix.

\subsection{Metric}\label{metric}

The leaderboard reports overall accuracy, the fraction of clips whose
mapped prediction matches the reference label. Class-level behavior is
summarized with precision and recall. For a class \(c\), let
\(\mathrm{TP}_c\), \(\mathrm{FP}_c\), and \(\mathrm{FN}_c\) denote true
positives, false positives, and false negatives:
\[
\mathrm{Precision}_c =
\frac{\mathrm{TP}_c}{\mathrm{TP}_c+\mathrm{FP}_c},
\qquad
\mathrm{Recall}_c =
\frac{\mathrm{TP}_c}{\mathrm{TP}_c+\mathrm{FN}_c}.
\]

Because each target class has 40 clips, per-class precision and recall
are also reported to expose class-specific failure modes and prediction
bias.

For aggregate accuracy, we also report a 95\% binomial confidence
interval. Each clip is treated as one Bernoulli trial, correct or
incorrect. The interval estimates the plausible range for the true
accuracy rate given the observed number of correct clips and total
scored clips, using a binomial proportion interval with finite-sample
correction. These intervals are descriptive. We do not claim
statistically significant ordering between models whose intervals
overlap substantially.

In addition to strict seven-way accuracy, we report a secondary
valence-bucket analysis. The valence view asks whether a model captures
a coarse affective direction even when it misses the exact emotion
label. This analysis maps
\texttt{angry}, \texttt{disgusted}, \texttt{fearful}, and \texttt{sad}
to negative, \texttt{happy} to positive, and \texttt{neutral} to
neutral, while excluding \texttt{surprised}.

\section{Baseline Models}\label{baseline-models}

The released benchmark includes six baselines spanning general
audio-capable models, speech-specific models, and provider
emotion/prosody endpoints. Table~\ref{tab:leaderboard} reports the
aggregate leaderboard.

\begin{table*}[t]\centering\scriptsize
  \caption{Seven-class VocalAffectBench baseline results. All baselines scored all 280 clips.}\label{tab:leaderboard}
  \begin{tabularx}{\textwidth}{LLrrrrrrrrr}
    \toprule
    Benchmark ID & Provider model & Correct & Acc. & angry & disgust. & fearful & happy & neutral & sad & surpr. \\
    \midrule
    \texttt{gemini\_3\_5\_flash} & \texttt{gemini-3.5-flash} & 124 & 44.3 & 57.5 & 27.5 & 32.5 & 45.0 & 75.0 & 70.0 & 2.5 \\
    \texttt{hume\_prosody} & \texttt{speech\_prosody} & 106 & 38.0 & 50.0 & 5.1 & 0.0 & 62.5 & 85.0 & 22.5 & 40.0 \\
    \texttt{tr\_qwen3\_5\_omni\_plus} & \texttt{qwen3.5-omni-plus} & 106 & 37.9 & 35.0 & 25.0 & 17.5 & 47.5 & 65.0 & 65.0 & 10.0 \\
    \texttt{tr\_voxtral\_small} & \texttt{mistralai/voxtral-small-24b-2507} & 96 & 34.3 & 17.5 & 72.5 & 17.5 & 35.0 & 47.5 & 27.5 & 22.5 \\
    \texttt{inworld\_voice\_profile} & \texttt{inworld/inworld-stt-1} & 80 & 28.6 & 37.5 & 0.0 & 12.5 & 30.0 & 97.5 & 17.5 & 5.0 \\
    \texttt{openai\_realtime} & \texttt{gpt-realtime-2} & 78 & 27.9 & 32.5 & 2.5 & 10.0 & 25.0 & 87.5 & 27.5 & 10.0 \\
    \bottomrule
  \end{tabularx}
\end{table*}

The strongest released baseline is \texttt{gemini\_3\_5\_flash}, with 124
correct predictions out of 280 and 44.3\% accuracy (95\% CI:
38.6--50.1). The lowest released baseline is
\texttt{openai\_realtime}, with 78 correct predictions and 27.9\%
accuracy (95\% CI: 22.9--33.4). Across all six baselines, there were 590 correct decisions, for an average accuracy
of 35.1\%.

\section{Results}\label{results}

\subsection{Class Difficulty}\label{class-difficulty}

Performance varies substantially across emotion classes.
Table~\ref{tab:class-difficulty} reports aggregate precision and recall
by label over the six released baselines.

\begin{table*}[t]\centering\footnotesize
  \caption{Aggregate per-class precision and recall across six baselines.}\label{tab:class-difficulty}
  \begin{tabular}{lrrrrr}
    \toprule
    Emotion & True & Pred. & Correct & Precision (\%) & Recall (\%) \\
    \midrule
    \texttt{neutral} & 240 & 761 & 183 & 24.0 & 76.2 \\
    \texttt{happy} & 240 & 247 & 98 & 39.7 & 40.8 \\
    \texttt{angry} & 240 & 149 & 92 & 61.7 & 38.3 \\
    \texttt{sad} & 240 & 247 & 92 & 37.2 & 38.3 \\
    \texttt{disgusted} & 240 & 129 & 53 & 41.1 & 22.1 \\
    \texttt{fearful} & 240 & 46 & 36 & 78.3 & 15.0 \\
    \texttt{surprised} & 240 & 100 & 36 & 36.0 & 15.0 \\
    \bottomrule
  \end{tabular}
\end{table*}

Precision and recall together help reveal label bias. \texttt{neutral}
has the highest recall, 76.2\%, but the lowest precision, 24.0\%,
because the baselines predict \texttt{neutral} 761 times across 1,679
decisions. That pattern indicates over-prediction of neutral rather than
uniformly reliable neutral recognition, while \texttt{fearful} shows the
opposite pattern, with 78.3\% precision but 15.0\% recall. Models rarely
predict \texttt{fearful}, but when they do the prediction is often
correct. These patterns show that model choice and safeguards should
depend on the emotion classes that matter most in a specific application.

\subsection{Neutral Bias}\label{neutral-bias-and-confusions}

The dominant error pattern is over-prediction of \texttt{neutral}. Across
all scored decisions, models predict \texttt{neutral} 761 times, or
45.3\% of all outputs. Among incorrect predictions, 578 of 1,089 errors are mapped to
\texttt{neutral}.

Table~\ref{tab:top-confusions} shows the most frequent aggregate
confusions. All six true non-neutral classes are often collapsed to
\texttt{neutral}. The largest confusion is \texttt{fearful} to
\texttt{neutral}, followed by \texttt{sad} to \texttt{neutral} and
\texttt{disgusted} to \texttt{neutral}. This matters for applications
because a conservative neutral prediction can make a voice product miss the
very affective states that should trigger escalation, empathy, or
additional caution.

\begin{table}[t]\centering\footnotesize
  \caption{Most frequent aggregate confusions across all baselines.}\label{tab:top-confusions}
  \begin{tabular}{lr}
    \toprule
    Confusion & Count \\
    \midrule
    \texttt{fearful} $\rightarrow$ \texttt{neutral} & 113 \\
    \texttt{sad} $\rightarrow$ \texttt{neutral} & 110 \\
    \texttt{disgusted} $\rightarrow$ \texttt{neutral} & 106 \\
    \texttt{happy} $\rightarrow$ \texttt{neutral} & 85 \\
    \texttt{surprised} $\rightarrow$ \texttt{neutral} & 85 \\
    \texttt{angry} $\rightarrow$ \texttt{neutral} & 79 \\
    \texttt{surprised} $\rightarrow$ \texttt{happy} & 58 \\
    \texttt{fearful} $\rightarrow$ \texttt{sad} & 47 \\
    \bottomrule
  \end{tabular}
\end{table}

\subsection{Valence as a Coarser Target}\label{valence-as-a-coarser-target}

Some products do not need a seven-way emotion label. They may only need
a coarser valence signal. As a secondary descriptive analysis, we map \texttt{angry},
\texttt{disgusted}, \texttt{fearful}, and \texttt{sad} to negative,
\texttt{happy} to positive, and \texttt{neutral} to neutral. We exclude
\texttt{surprised} because its valence is ambiguous.

Under this coarse mapping, aggregate accuracy rises to 49.2\%. The best model on this valence view is
\texttt{gemini\_3\_5\_flash} at 66.2\%, followed by
\texttt{tr\_voxtral\_small} at 55.0\% and
\texttt{tr\_qwen3\_5\_omni\_plus} at 52.5\%. This suggests that valence
is more tractable than discrete emotion for current models, but still
not solved.

Accuracy alone does not show whether a model tends to emit positive,
neutral, or negative labels. We therefore also describe each model's
output valence skew in Table~\ref{tab:valence-skew}. This should not be
read as an intrinsic property of the model. It is a descriptive summary
of the labels the model outputs under this benchmark and mapping.

\begin{table*}[t]\centering\footnotesize
  \caption{Output valence skew based on mapped predictions. Percentages are shares of all 280 model outputs.}\label{tab:valence-skew}
  \begin{tabular}{lrrrrl}
    \toprule
    Model & Neg. out. & Pos. out. & Neutral out. & Ambig. out. & Output skew \\
    \midrule
    \texttt{gemini\_3\_5\_flash} & 51.4 & 10.7 & 36.8 & 1.1 & Negative \\
    \texttt{hume\_prosody} & 24.4 & 27.6 & 36.2 & 11.8 & Neutral \\
    \texttt{tr\_qwen3\_5\_omni\_plus} & 42.9 & 12.9 & 41.8 & 2.5 & Negative \\
    \texttt{tr\_voxtral\_small} & 50.7 & 16.4 & 20.0 & 12.9 & Negative \\
    \texttt{inworld\_voice\_profile} & 16.8 & 14.6 & 67.5 & 1.1 & Neutral \\
    \texttt{openai\_realtime} & 17.9 & 6.1 & 69.6 & 6.4 & Neutral \\
    \bottomrule
  \end{tabular}
\end{table*}

By this definition, three baselines are negative-skewed and three are
neutral-skewed. None of the evaluated baselines is positive-skewed. This
matters because two models with similar aggregate accuracy can create
different product risks if one over-predicts negative affect and another
over-predicts neutral delivery.

\section{Discussion}\label{discussion}

The baseline results show partial but brittle affect recognition. The
leading model is roughly three times the random baseline, but still
misses more than half of the clips. Average performance across models is
only 35.1\%. These results make the benchmark useful as a stress test:
it exposes the gap between detecting some vocal affect and relying on a
seven-way emotion label in production.

The class-level pattern is as important as the aggregate ranking.
Models performed best on \texttt{neutral}, while \texttt{surprised},
\texttt{fearful}, and \texttt{disgusted} remain difficult for most
baselines. This creates a practical risk. A model may appear useful when
averaged over a neutral-heavy workload, yet fail on the high-salience
states that matter most for escalation or intervention.

The neutral-bias result also cautions against treating speech emotion
outputs as facts about users. A neutral prediction can mean that the
model did not detect salient affect, but it can also reflect model
uncertainty or calibration error. Deployed products should therefore be
conservative when affect signals would change material outcomes.

For voice agents, emotion recognition should be treated as a
probabilistic auxiliary signal after product-specific validation, not as
a sole controller of action. The current baselines are better viewed as
measurement targets than as deployable decision mechanisms.

\section{Intended Use}\label{intended-use}

VocalAffectBench is intended for diagnostic evaluation of audio emotion
recognition models. Appropriate uses include provider comparison, error
analysis, regression tracking, and class-specific product validation.

The benchmark is not intended as a training corpus, a hidden leaderboard,
a universal measure of emotional intelligence, or a biometric dataset.
Since labels and baseline predictions are public, reports should include
enough evaluation detail to reproduce the result.

\section{Limitations}\label{limitations}

VocalAffectBench is intentionally focused. First, it is English-only and
all released clips are spoken in a General American accent. Emotion 
expression, prosody, speech
rhythm, and label interpretation vary across languages and cultures.
Multilingual and cross-accent evaluation will require additional
collection and label validation rather than direct translation.

Second, the benchmark uses a seven-label discrete taxonomy. Human affect
is richer than a single class. A primary label is useful for
reproducible benchmarking, but it cannot capture every state a listener
might perceive. The included valence analysis partly addresses this by
showing a coarser alternative, but it is not a replacement for richer
affect modeling.

Third, clips were selected to give each target class the same number of
examples. This is useful for fair comparison and per-class analysis, but
it is not a natural estimate of emotion prevalence in production
conversations.
Models used in real workloads should be evaluated against their own
traffic distribution as well.

Fourth, the dataset contains acted or performed emotional speech. This
matches the benchmark target of expressed vocal emotion, but it limits
what the results can claim about spontaneous real-world conversations.
Results should not be interpreted as evidence that a model can infer
what a speaker truly feels.

Fifth, the reference labels are assigned performance targets verified by
one reviewer. This provides a practical benchmark signal, but it does
not measure population-level listener perception and does not support
inter-rater reliability analysis.

\section{Ethical and Privacy Considerations}\label{ethical-and-privacy-considerations}

Released speech can contain acoustic characteristics that may identify
or profile speakers, even when the text content is not sensitive.
Contributors consented to public release of their recordings for
research and benchmarking use. The dataset is nevertheless released for
evaluation, not for biometric or high-stakes use.

The benchmark labels expressed performance rather than inner state. This
framing should be preserved in downstream reporting. Models evaluated
on VocalAffectBench should avoid claims that they can determine a
person's inner state or personal risk level from voice alone.

\section{Data Availability}\label{data-availability}

VocalAffectBench is a public, test-only benchmark. The
dataset and evaluation harness use the MIT License, the
same release license used for Voice Code Bench. Because the dataset
contains human voice recordings, users should still follow the
intended-use and ethics constraints in this paper and the dataset card.
The dataset repository is available at
\url{https://huggingface.co/datasets/besimple-ai/vocal-affect-bench}.

The release includes the audio, metadata, predictions, aggregate
results, and documentation. The main metadata file is
\texttt{data/metadata.jsonl}. Baseline predictions are stored in
\texttt{data/predictions.csv} and \texttt{baselines/predictions/}.
Aggregate results are stored in \texttt{data/leaderboard-summary.csv}
and \texttt{baselines/results.csv}. The public release does not include
source materials or demographic metadata.

\section{Conclusion}\label{conclusion}

VocalAffectBench focuses attention on a practical voice-agent
requirement that transcript-based evaluation can obscure: audio models
must preserve affective cues carried by vocal delivery, not only the
words being spoken. Its contribution is not another training corpus for
speech emotion recognition, but a test of whether raw-audio models can
recover expressed vocal emotion under a fixed, auditable protocol. By
combining controlled label construction, transcript-free evaluation,
released predictions, and per-class error analysis, the benchmark
connects emotion recognition performance to product risk.

The baseline results show why this distinction matters. Current audio
models extract some affective signal, but seven-way expressed-emotion
recognition remains brittle; neutral predictions are overused; and
several non-neutral classes are frequently missed. Reporting aggregate
accuracy alongside per-class precision, recall, confusions, and valence
skew therefore gives model developers and application teams a more
actionable view of quality: which models detect broad affective signal,
which miss high-salience emotions, and where additional validation or
safeguards are needed before affect predictions influence user-facing
behavior.

\section{References}\label{references}

Busso, C., Bulut, M., Lee, C.-C., Kazemzadeh, A., Mower, E.,
Kim, S., Chang, J. N., Lee, S., and Narayanan, S. S. 2008. IEMOCAP:
Interactive Emotional Dyadic Motion Capture Database. \emph{Language
Resources and Evaluation}, 42(4):335--359.

Cao, H., Cooper, D. G., Keutmann, M. K., Gur, R. C., Nenkova, A., and
Verma, R. 2014. CREMA-D: Crowd-Sourced Emotional Multimodal Actors
Dataset. \emph{IEEE Transactions on Affective Computing}, 5(4):377--390.

Ekman, P. 1992. An Argument for Basic Emotions. \emph{Cognition and
Emotion}, 6(3--4):169--200.

Eyben, F., Scherer, K. R., Schuller, B. W., Sundberg, J.,
Andre, E., Busso, C., Devillers, L. Y., Epps, J., Laukka, P.,
Narayanan, S. S., and Truong, K. P. 2016. The Geneva Minimalistic
Acoustic Parameter Set (GeMAPS) for Voice Research and Affective
Computing. \emph{IEEE Transactions on Affective Computing}, 7(2):190--202.

Livingstone, S. R., and Russo, F. A. 2018. The Ryerson Audio-Visual
Database of Emotional Speech and Song (RAVDESS): A Dynamic, Multimodal
Set of Facial and Vocal Expressions in North American English.
\emph{PLOS ONE}, 13(5):e0196391.

Russell, J. A. 1980. A Circumplex Model of Affect. \emph{Journal of
Personality and Social Psychology}, 39(6):1161--1178.

Schuller, B., Steidl, S., Batliner, A., Bergelson, E., Krajewski, J.,
Janott, C., Amatuni, A., Casillas, M., Seidl, A., Soderstrom, M.,
Warlaumont, A., Hidalgo, G., Schnieder, S., Heiser, C., Hohenhorst, W.,
Herzog, M., Schmitt, M., Qian, K., Zhang, Y., and Zafeiriou, S. 2018.
The INTERSPEECH 2018 Computational Paralinguistics Challenge:
Atypical and Self-Assessed Affect, Crying and Heart Beats. In
\emph{Proceedings of Interspeech 2018}, 122--126.

\appendix

\section{Appendix: Released File Schema}\label{appendix-file-schema}

Each row in \texttt{data/metadata.jsonl} is a JSON object with the
following fields:
\begin{verbatim}
{
  "audio_id": "06Eg9dXO99fAAti4HB34",
  "file_name": "audio/06Eg9dXO99fAAti4HB34.wav",
  "required_emotion": "neutral",
  "duration_seconds": 12.4,
  "sample_rate": 16000,
  "channels": 1
}
\end{verbatim}

The main prediction file uses the following CSV columns:
\begin{verbatim}
audio_id,required_emotion,model_name,provider_model,
predicted_label,mapped_label,confidence,correct,error
\end{verbatim}

\section{Appendix: Valence View}\label{appendix-valence-view}

Table~\ref{tab:valence} reports the secondary valence analysis described
in the section ``Valence as a Coarser Target.'' The mapping is:
\texttt{angry}, \texttt{disgusted}, \texttt{fearful}, and \texttt{sad}
map to negative. \texttt{happy} maps to positive. \texttt{neutral} maps
to neutral.
\texttt{surprised} is excluded.

\begin{table}[!htbp]\centering\footnotesize
  \caption{Secondary valence accuracy, excluding \texttt{surprised}.}\label{tab:valence}
  \begin{tabular}{lrrr}
    \toprule
    Model & Scored & Correct & Accuracy \\
    \midrule
    \texttt{gemini\_3\_5\_flash} & 240 & 159 & 66.2 \\
    \texttt{tr\_voxtral\_small} & 240 & 132 & 55.0 \\
    \texttt{tr\_qwen3\_5\_omni\_plus} & 240 & 126 & 52.5 \\
    \texttt{hume\_prosody} & 239 & 114 & 47.7 \\
    \texttt{inworld\_voice\_profile} & 240 & 92 & 38.3 \\
    \texttt{openai\_realtime} & 240 & 85 & 35.4 \\
    \midrule
    Aggregate & 1439 & 708 & 49.2 \\
    \bottomrule
  \end{tabular}
\end{table}

\section{Appendix: Exact Prompt}\label{appendix-exact-prompt}

Prompted audio-model baselines use the following instruction:

\begin{verbatim}
You are evaluating vocal expression in an audio clip.

Choose exactly one primary expressed emotion from this allowed label set:
[angry, disgusted, fearful, happy, neutral, sad, surprised]

Use one of those labels verbatim. Do not use synonyms or any emotion
outside the allowed label set.

Do not infer the speaker's private internal state.
Base your answer only on the expressed vocal tone, prosody, pace,
intensity, pauses, and wording.

Return only valid JSON:
{
  "primary_emotion": "...",
  "confidence": 0.0,
  "evidence": "brief explanation"
}
\end{verbatim}

\section{Appendix: Label Mapping}\label{appendix-label-mapping}

Provider-native labels and prompted-model synonyms are normalized to the
seven-label benchmark set before scoring. The Hume-specific mapping is:

\begin{verbatim}
aesthetic appreciation -> happy
amusement -> happy
anger -> angry
anxiety -> fearful
boredom -> neutral
calmness -> neutral
concentration -> neutral
confusion -> surprised
contemplation -> sad
contentment -> happy
disappointment -> sad
determination -> angry
disgust -> disgusted
distress -> sad
excitement -> happy
fear -> fearful
horror -> fearful
interest -> surprised
joy -> happy
pride -> happy
realization -> surprised
sadness -> sad
satisfaction -> happy
surprise (negative) -> surprised
surprise (positive) -> surprised
tiredness -> neutral
\end{verbatim}

The general alias mapping used for prompted and other provider outputs
is:

\begin{verbatim}
anger/angry -> angry
anxious/anxiety/fear/fearful/scared -> fearful
bored/boredom/calm/calmness/concentration/tender/tiredness -> neutral
contemplation/sad/sadness -> sad
confused/confusion/interest/realization/surprise/surprised -> surprised
determination/frustrated/frustration -> angry
disgust/disgusted -> disgusted
excited/excitement/happy/happiness/joy/joyful/pride/satisfaction -> happy
unclear -> unscored
\end{verbatim}

\section{Appendix: Citation}\label{appendix-citation}

\begin{verbatim}
@misc{vocalaffectbench2026,
  title  = {VocalAffectBench: Evaluating Vocal Emotion Recognition in AI Audio Models},
  author = {Debaupte, Luc and Baumgartner, Tyler and Tai, Brandon and Fan, Candice and Wang, Bill and Zhong, Yi},
  year   = {2026},
  note   = {Benchmark dataset},
  url    = {https://huggingface.co/datasets/besimple-ai/vocal-affect-bench},
  license = {MIT}
}
\end{verbatim}

\end{document}